\documentclass[letterpaper, 10 pt, conference]{ieeeconf}  

\usepackage{amsmath,amssymb,amsfonts}
\usepackage{balance}
\usepackage{bm}
\usepackage{cite}
\usepackage{hyperref}
\usepackage{color}
\usepackage[table,xcdraw]{xcolor}
\usepackage{tabularx}
\usepackage{booktabs}
\usepackage{graphicx}
\usepackage{multirow}
\usepackage{tikz}
\usepackage{pgfplots}
\usepackage{units}
\usetikzlibrary{bending, external, plotmarks}

\IEEEoverridecommandlockouts                             
\pgfplotsset{compat=1.16} 

\newcommand{\wfr}[0]{\ensuremath{\mathcal{W}}} %
\newcommand{\bfr}[0]{\ensuremath{\mathcal{B}}} %

\newcolumntype{C}{>{\centering\arraybackslash}X}

\definecolor{somegray}{rgb}{0.5, 0.5, 0.5}
\newcommand{\darkgrayed}[1]{\textcolor{somegray}{#1}}
\makeatletter
\newcommand*\titleheader[1]{\gdef\@titleheader{#1}}
\AtBeginDocument{%
  \let\st@red@title\@title
  \def\@title{%
    \vskip-3em
    \bgroup\normalfont\large\centering\@titleheader\par\egroup
    \vskip1.1em\st@red@title}
}
\makeatother

\titleheader{\darkgrayed{This paper has been accepted for publication at the \\ IEEE International Conference on Robotics and Automation (ICRA), London, 2023. \copyright IEEE}}

\title{\LARGE \bf
User-Conditioned Neural Control Policies for Mobile Robotics
}

\author{Leonard Bauersfeld, Elia Kaufmann, Davide Scaramuzza
    \thanks{
    The authors are with the Robotics and Perception Group, Department of Informatics, University of Zurich, and Department of Neuroinformatics, University of Zurich and ETH Zurich, Switzerland (\protect\url{http://rpg.ifi.uzh.ch}). 
    This work was supported by the Swiss National Science Foundation (SNSF) through the National Centre of Competence in Research (NCCR) Robotics, the European Union’s Horizon 2020 Research and Innovation Programme under grant agreement No. 871479 (AERIAL-CORE), and the European Research Council (ERC) under grant agreement No. 864042 (AGILEFLIGHT).
    }%
}

\begin{document}

\maketitle
\thispagestyle{empty}
\pagestyle{empty}

\begin{abstract}
Recently, learning-based controllers have been shown to push mobile robotic systems to their limits and provide the robustness needed for many real-world applications.
However, only classical optimization-based control frameworks offer the inherent flexibility to be dynamically adjusted during execution by, for example, setting target speeds or actuator limits. 
We present a framework to overcome this shortcoming of neural controllers by conditioning them on an auxiliary input. This advance is enabled by including a feature-wise linear modulation layer~(FiLM). 
We use model-free reinforcement-learning to train quadrotor control policies for the task of navigating through a sequence of waypoints in minimum time.
By conditioning the policy on the maximum available thrust or the viewing direction relative to the next waypoint, a user can regulate the aggressiveness of the quadrotor's flight during deployment. 
We demonstrate in simulation and in real-world experiments that a single control policy can achieve close to time-optimal flight performance across the entire performance envelope of the robot, reaching up to 60~km/h and 4.5\,g in acceleration.
The ability to guide a learned controller during task execution has implications beyond agile quadrotor flight, as conditioning the control policy on human intent helps safely bringing learning based systems out of the well-defined laboratory environment into the wild.
\end{abstract}

\begin{center}
\small
Video: \url{https://youtu.be/rwT2QQZEH6U}
\end{center}
\vspace*{-6pt}

\vspace*{-3pt}
\section{Introduction}
\vspace*{-3pt}
Recently, learned controllers have become extremely popular in the mobile robotics community due to their success in a variety of complex real-world tasks, such as legged locomotion in challenging environments~\cite{lee2020animalroughterrain}, underground exploration~\cite{tranzatto2021cerebus}, autonomous drone racing~\cite{foehn2020alphapilot, ackermann2022ieee, song2021autonomous}, and virtual car racing~\cite{wurman2022gtsophy}. 
In all the aforementioned works, neural-network controllers outperform their classical model-based counterparts both in terms of performance and success rate. 
However, this performance comes at the expense of adaptability, as the control approaches are trained to overfit on a narrowly defined task.
A standard neural controller can only rigidly execute the specific task that it has been trained on and lacks the versatility of traditional model-based control.
Consider, for example, a mobile robot tasked with time-optimal navigation: using model-predictive control~(MPC) it would be straightforward to limit the maximum acceleration during deployment by adjusting the actuator constraints inside the model~\cite{romero2022mpcc}. However, most neural controllers cannot be regulated and naively adding an additional input to the learned policy may not lead to the desired performance. 

\begin{figure}[t!]
    \centering
    \includegraphics[]{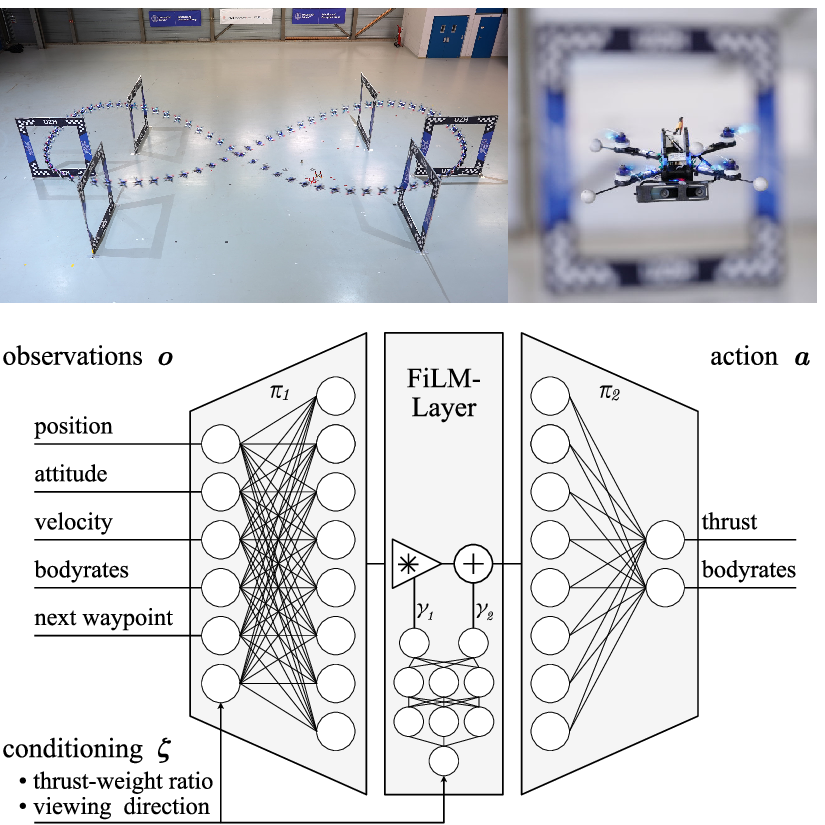}
    \vspace*{-6pt}
    \caption{Conditioning a control policy for agile quadrotor flight on an auxilliary input can be achieved through a FiLM architecture \cite{perez2018film}. There, the intermediate activations of a policy that directly maps observations to control commands are linearly transformed based on the conditioning signal supplied by the user. In this work, we study conditioning on the maximum thrust-to-weight ratio (agility) and the viewing direction of the drone w.r.t the next waypoint.} 
    \label{fig:figure_1}
    \vspace*{-24pt}
\end{figure}

This paper proposes an approach to alleviate the drawback of rigid task execution of learning-based controllers by conditioning the control policies on an auxiliary input which an operator (human, high-level planner) can then set to influence the neural controller as shown in Fig.~\ref{fig:figure_1}. 
Aside from increasing the versatility, training a policy that can not only react to the environment but also condition its computed control actions on human intent allows safely bringing learning-based systems out of the controlled laboratory environment into the wild. 

\begin{figure*}[t!]
\includegraphics[]{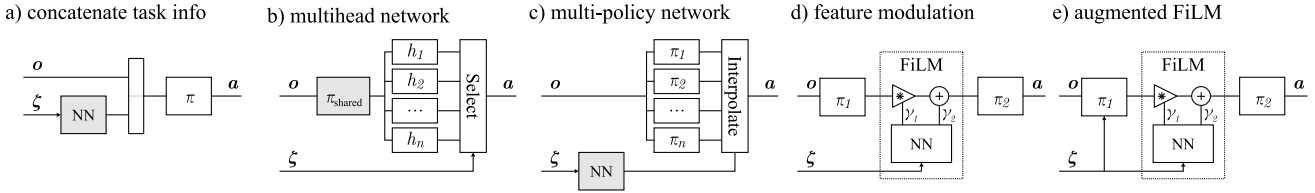}
\vspace*{-12pt}
\caption{Overview of the different architectures for conditioning of neural networks commonly found in literature. Boxes in gray represent optional components that can also be replaced with a direct connection. The variable $\bm{o}$ denotes some vector of observations (e.g. the system state) supplied to the policy. The conditioning signal is denoted by $\bm{\zeta}$ and the output of the network (e.g. control action) is denoted by $\bm{a}$.}
\label{fig:architectures}
\vspace*{-18pt}
\end{figure*}

Training an embodied agent that can react to user inputs is a difficult endeavor as it requires to learn an entire \emph{distribution} of policies, as opposed to learning a static policy that maps sensory observations to actions.
Prior work primarily exists in the context of robotic manipulation conditioned on visual or natural language input~\cite{mees2022languageconditionedlearning, lynch2020languageconditioned}.
This so-called \emph{multi skill learning} strongly focuses on handling complex visual or natural language queries~\cite{reed2022gato} while the robotic system is mostly simulated and has minimal complexity from a control engineering perspective.
Closely related are the visual question answering tasks encountered in the computer vision community, where networks are again conditioned on complex natural language user queries~\cite{perez2018film}.
In the context of mobile robotics, to the best of the authors' knowledge, only one prior work~\cite{codevilla2018endtoenddriving} exists where conditioning has been applied for three discrete user inputs: 
a remote-controlled car is trained to either turn left, go straight, or turn right at intersections by using a control network with a shared encoder and three disjoint network heads that are selected based on the operator's input.
\vspace*{-2pt}
\subsection{Contribution}
\vspace*{-2pt}
We present the first learning-based controller for an autonomous mobile robot\textemdash an agile quadrotor platform is used in this work\textemdash where the vehicle's agility and its viewing direction can be influenced through a continuous conditioning input supplied by a user.
This advance is made possible by integrating a modified version of the feature-wise linear modulation layer (FiLM)~\cite{perez2018film} into the neural network controller, which is trained using model-free reinforcement learning. 
To support our choice of the FiLM architecture, we present a large ablation study which compares the commonly used methods in multi-skill learning tasks. 
The FiLM approach outperforms multi-head networks as well as a naive feature concatenation baseline in terms of performance and robustness. 
Finally, we demonstrate the applicability of our proposed method to real-world mobile robotics by conditioning a control policy for perception-aware, near time-optimal quadrotor flight. The continuous user input regulates the desired agility level or guides a perception objective. 
Furthermore we show that there is a less than a 2\% performance difference between a single policy conditioned on a user-specified agility level and a set of overfit policies that can only operate at a fixed level.

\vspace*{-2pt}
\section{Related Work}
\vspace*{-2pt}
Outside the field of mobile robotics, conditioning neural networks on auxiliary user inputs has been studied in recent years for a variety of applications. 
In most of them, the conditioning signal is given by a natural language prompt specified by the user and the conditioned network is tasked with answering a question~\cite{perez2018film}, controlling a robotic manipulator arm~\cite{mees2022languageconditionedlearning, lynch2020languageconditioned} in a specified way, or planning a path such that a vehicle visits specific areas on a map~\cite{blukis2018mappingnavigation}. 
Other tasks studied in literature range from optimally encoding information~\cite{dosovitskiy2020yoto} to throwing simulated darts at different targets~\cite{silva2021learningparametrizedskills}. Yet, the only application to mobile robotics is driving a remote-controlled car autonomously~\cite{codevilla2018endtoenddriving} and conditioning on the direction it needs to turn at an intersection. 
However, in the context of this work, it is more informative to compare the works in terms of the architectures they leverage to condition their networks. Figure~\ref{fig:architectures} presents a summary of the common approaches found in literature.

The conceptually simplest approach to conditioning neural networks is shown in Fig.~\ref{fig:architectures}\,(a) where the conditioning signal $\boldsymbol\zeta$ is simply appended to the policy observation $\bm o$~\cite{deisenroth2014policysearch}. 
As such approach is only possible with continuous scalar/vector inputs, many works include an encoding network~\cite{mees2022languageconditionedlearning, dosovitskiy2017learning, blukis2018mappingnavigation, Abramson2021Deepmindmultimodal, rahmatizadeh2018visionbasedmultitask}, which generates a numeric representation of the conditioning signal. 
Such encoders can be represented by fully connected layers~\cite{dosovitskiy2017learning}, transformers~\cite{mees2022languageconditionedlearning}, recurrent neural networks~\cite{perez2018film} for natural language conditioning signals, or convolutional networks for image-based conditioning~\cite{blukis2018mappingnavigation}.

The multihead (b) and multipolicy (c) architectures shown in Fig.~\ref{fig:architectures} are very similar.
In a multihead network all heads operate on the same latent representation produced by a shared encoder $\pi_\text{shared}$. 
A subsequent multiplexer then selects one of the heads based on the current task signal.
This architecture has been applied successfully to a real-world remote-controlled car, which can either turn left, go straight, or turn right at intersections based on the conditioning signal $\boldsymbol\zeta$~\cite{codevilla2018endtoenddriving}. 
In this form the approach can only be applied when the task-space is discrete and a head for each discrete task-space class exists, e.g. three heads are required for a controller that enables the car to go left, straight, right.
The multi-policy approach with a subsequent interpolation layer presented in~\cite{silva2021learningparametrizedskills, schaul2015universalvalue} is very similar.
However, no shared encoder is used and the multiplexer is replaced by an interpolation module.
The latter enables this approach to handle continuous task signals as the individual control actions by the respective policies are combined smoothly. 

A novel approach to conditioning a network that does not require training separate heads nor performs naive input feature concatenation presented in~\cite{perez2018film}.
Their proposed feature-wise linear modulation (FiLM) layer is illustrated in Fig.~\ref{fig:architectures}\,d). 
The idea is that a FiLM layer is inserted between two layers of an existing network, effectively splitting the original network into two parts $\pi_1$ and $\pi_2$. 
The activations of the first part $\pi_1$ are passed through the FiLM layer which applies an affine transform with trainable parameters $\gamma_1$ and $\gamma_2$. 
The transformed activations are then used as input to $\pi_2$ which generates the final control action. 
This architecture was originally devised for transforming feature maps of convolutional neural networks~\cite{perez2018film} but has been applied to robotic manipulation tasks~\cite{jang2021bcz}, optimal information encoding, and style transfer tasks~\cite{dosovitskiy2020yoto}. 
As an extension, we also propose an augmented FiLM architecture Fig.~\ref{fig:architectures}~e) which also feeds the conditioning signal into the control policy directly.

\vspace*{-2pt}
\section{Methodology}
\vspace*{-2pt}
In this work we will compare and evaluate the different architectures shown in Fig.~\ref{fig:architectures} for the task of conditioning a quadrotor control on user input.
Focusing on the challenging task of agile quadrotor flight, policies are trained using model-free reinforcement learning and directly map a set of observations~$\bm{o}_t$ to low-level control actions~$\bm{a}_t$~\cite{kaufmann2022benchmark}.
This section first presents a brief overview of the quadrotor simulator used for training, then proceeds to explain the neural controller, and concludes by introducing the demonstrators evaluated in this work.

\vspace*{-2pt}
\subsection{Notation \& Quadrotor Dynamics}
\vspace*{-2pt}
Throughout this paper, scalars are denoted in non-bold~$[s, S]$, vectors in lowercase bold~$\bm{v}$, and matrices in uppercase bold~$\bm{M}$.
World $\wfr$ and Body $\bfr$ frames are defined with an orthonormal basis i.e. $\{\bm{x}_\wfr, \bm{y}_\wfr, \bm{z}_\wfr\}$.
The frame $\bfr$ is located at the center of mass of the quadrotor.

The quadrotor is assumed to be a 6 degree-of-freedom rigid body of mass $m$ and diagonal moment of inertia matrix $\bm{J}=\mathrm{diag}(J_x, J_y, J_z)$.
Furthermore, the rotational speeds of the four propellers $\Omega_i$ are modeled as a first-order system with time constant $k_\text{mot}$ where the commanded motor speeds $\boldsymbol\Omega_\text{cmd}$ are the input.
The state space is thus 17-dimensional and its dynamics can be written as:
\begin{align}
\renewcommand{\arraystretch}{1.6}
\small
\label{eq:3d_quad_dynamics}
\dot{\bm{x}} =
\begin{bmatrix}
\dot{\bm{p}}_{\wfr\bfr} \\
\dot{\bm{q}}_{\wfr\bfr} \\
\dot{\bm{v}}_{\wfr} \\
\dot{\boldsymbol\omega}_\bfr \\
\dot{\boldsymbol\Omega}
\end{bmatrix} = 
\begin{bmatrix}
\bm{v}_\wfr \\
\bm{q}_{\wfr\bfr} \cdot \begin{bmatrix}0 & \bm{\omega}_\bfr/2\end{bmatrix}^\top \\
\frac{1}{m} \left( \bm{q}_{\wfr\bfr} \odot \left(\bm{f}_\text{prop} + \bm{f}_\text{res}\right)\right) +\bm{g}_\wfr  \\
\bm{J}^{-1}\big(\boldsymbol{\tau}_\text{prop} + \boldsymbol\tau_\text{res} - \boldsymbol\omega_\bfr \times \bm{J}\boldsymbol\omega_\bfr \big) \\
\frac{1}{k_\text{mot}} \big(\boldsymbol\Omega_\text{cmd} - \boldsymbol\Omega \big)
\end{bmatrix} \; ,
\end{align}
where $\bm{g}_\wfr= [0, 0, \unit[-9.81]{ms^{-2}}]^\top$ denotes earth's gravity, $\bm{f}_\text{prop}$, $\boldsymbol{\tau}_\text{prop}$ are the collective force and the torque produced by the propellers. To account for residual aerodynamic effects, we introduce a lumped residual term $\bm{f}_\text{res}$, $\boldsymbol\tau_\text{res}$ on the forces and torques respectively.

Model-free reinforcement learning suffers from a low sample-efficiency during training which necessitates an efficient simulator that can run fast. 
Hence, to model the thrust and torque produced by the $i$-th propeller, the commonly used and computationally lightweight quadratic model \cite{shah2018airsim, furrer2016rotors, foehn2022agilicious} is employed:
\begin{align}
    \bm{f}_i(\Omega) &= \begin{bmatrix} 0 ~~ 0 ~~ c_{\text{l}}\,\Omega^2  \end{bmatrix}^\top &
    \bm{\tau}_i(\Omega) &= \left[0 ~~ 0 ~~ c_{\text{d}}\,\Omega^2\right]^\top \\
    \bm{f}_\text{prop} &= \sum_i \bm{f}_i &
   \boldsymbol{\tau}_\text{prop} &= \sum_i \boldsymbol{\tau}_i + \bm{r}_{\text{P},i} \times \bm{f}_i 
\end{align}
where $c_\text{l}$, $c_\text{d}$ denote the lift and drag coefficient of propeller respectively and $\bm r_\text{P}$. 
Compared to state-of-the-art methods that leverage blade-element-momentum theory~\cite{bauersfeld2021neurobem}, this quadratic model does not account for aerodynamic effects, such as rotor drag or blade flapping. 
This deficiency widens the sim-to-real gap when deploying the trained controller in the real-world. 
To increase the simulation fidelity while maintaining low computational complexity we use a polynomial graybox model~\cite{sun2019graybox, bauersfeld2021neurobem} for the residual force $\bm f_\text{res}$ and torque $\boldsymbol\tau_\text{res}$ term.

\subsection{Neural Controller}
In this work, the task of fast and agile quadrotor flight is defined as navigating through a sequence of drone racing gates as fast as possible.
Or, to rephrase this using broader terms: Navigate through a sequence of predefined waypoints $g_i$  in minimum time and pass each waypoint within an $l_\infty$ distance less then the dimension of a racing gate.
To accomplish this, the control policy directly maps an observation $\bm o_t$ and a conditioning input $\zeta_t$ to an action (control command) $\bm a_t$. 
The control policies are trained using model-free reinforcement learning (PPO~\cite{schulman2017ppo}) purely in simulation. 

\subsubsection{Observation and Action Space}
At each timestep~$t$ the policy has access to an observation $\bm o_t$ from the environment which contains 
(i)~the current robot state, 
(ii)~the relative position to the next waypoint to be passed,
and (iii)~the current conditioning signal. 
Specifically, the state consists of the vehicle position $\bm p_{\wfr\bfr}$, its velocity in body-frame $\bm v_\bfr$ and its attitude. 
To avoid discontinuities the latter is represented by a rotation matrix instead of directly using the quaternion $\bm q_{\wfr\bfr}$~\cite{zhou2019continuity}.
The value network used during training time has access to the same input features as the policy network. In contrast to the policy network, the value network architecture does not contain any FiLM layers.

The control command $\bm a_t$ consists of a desired collective mass-normalized thrust $c$ and a bodyrate setpoint $\boldsymbol\omega_{\bfr, \text{ref}}$. 
Those commands are then tracked by a low-level controller, which finally controls the motors. 
In contrast to more abstract control modalities such as linear velocity references, operating on collective thrust and bodyrates has been shown to be well suited for agile learned quadrotor control~\cite{kaufmann2022benchmark}.

\subsubsection{Conditioning}
\label{sec:conditioning}
We compare and evaluate different network architectures (illustrated in Fig.~ \ref{fig:architectures}) to condition a neural controller for agile quadrotor flight. 
Specifically, the following architectures are considered: 
\begin{itemize}
    \item \emph{Naive-c} a naive baseline (see Fig.~\ref{fig:architectures}~a)) where continuous scalar conditioning signal is concatenated with the observation,
    \item \emph{Naive-d} the same architecture as \emph{naive-c} but with a discretized one-hot vector encoding of the conditioning signal, 
    \item \emph{Multihead} an architecture (see Fig.~\ref{fig:architectures}~b)) with a discrete conditioning signal similar to~\cite{codevilla2018endtoenddriving},
    \item \emph{FiLM-c} a standard FiLM architecture (see Fig.~\ref{fig:architectures}~d)) with a continuous scalar conditioning input,
    \item \emph{FiLM*-c} our augmented FiLM architecture (see Fig.~\ref{fig:architectures}~e)) with a continuous scalar conditioning input,
    \item \emph{FiLM*-d} the same architecture as \emph{FiLM*-c} but with a discretized one-hot vector encoding of the conditioning signal. 
\end{itemize} 

\begin{figure*}[t!]
    \centering
    \includegraphics[]{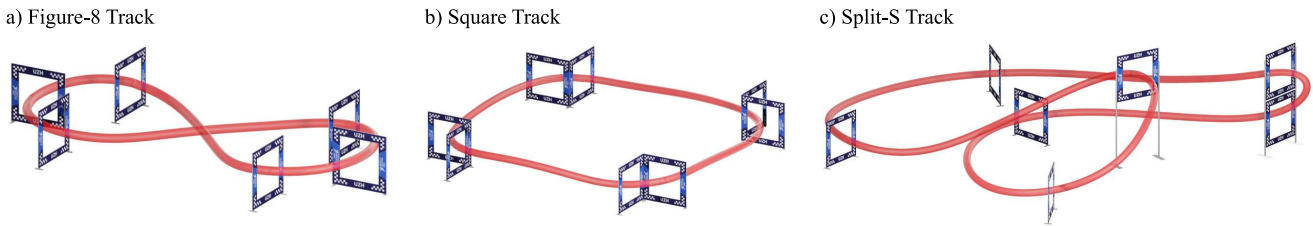}
    \vspace*{-6pt}
    \caption{We evaluate neural policy conditioning on the task of autonomous drone racing. Different approaches for policy conditioning are evaluated on a set of three different race tracks of varying complexity. 
    All tracks are of similar size, spanning between \unit[10]{m} and \unit[16]{m} in width.}
    \label{fig:racetracks}
    \vspace*{-12pt}
\end{figure*}

\subsubsection{Reward Function}
We use a dense shaped reward to encode the task of high-speed flight through a set of pre-defined waypoints. 
The reward $r_t$ at time step $t$ is given by 
\begin{align}
        r_t = & r_t^\text{prog} + r_t^\text{perc}(\zeta) - r_t^\text{twr}(\zeta) - r_t^\text{crash}\; ,
\end{align}
where $r^\text{prog}$ rewards progress towards the next gate to be passed~\cite{song2021autonomous}, $r^\text{perc}(\zeta)$ encodes perception awareness by adjusting the vehicle's attitude such that the optical axis of its camera points towards the next gate's center with an optional user-specified offset, $r^\text{twr}(\zeta)$ is a penalty for violating the user-specified maximum thrust-to-weight ratio, and $r^\text{crash}$ is a binary penalty that is only active when colliding with a gate or when the platform leaves a pre-defined bounding box, which also ends the episode.

Progress, perception, thrust-to-weight, and collision reward components are formulated as follows:
\begin{align}
    r_t^\text{prog} &= \lambda_1 \left( d_\text{Gate}(t-1) - d_\text{Gate}(t) \right) \nonumber\\
    r_t^\text{perc}(\zeta) &= \lambda_2 \exp \left(\lambda_3 \cdot \delta_\text{cam}(\zeta)^4\right) \\
    r_t^\text{twr}(\zeta) &= \max(\lambda_4 \exp\left( \lambda_5 (c_\text{cmd} - c_\text{twr}(\zeta)) \;/\; c_\text{max} \right) - 1, 0) \nonumber \\
    r_t^\text{crash} &= \begin{cases}
                  -5.0, & \text{if $p_z<0$ or in collision with gate}.\\
                  0, & \text{otherwise}
  \end{cases}\nonumber \; ,
\end{align}
where $d_\text{Gate}(t)$ denotes the distance from the quadrotor's center of mass to the center of the next gate, $\delta_\text{cam}(\zeta)$ is the angle between the optical axis of the camera and the user-specified viewing direction (center of the next gate + offset angle). The parameters $c_\text{cmd}$, $c_\text{twr}(\zeta)$ and $c_\text{max}$ are the commanded mass normalized thrust, the current user-specified maximum allowable mass normalized thrust and the maximum mass normalized thrust physically available for the quadrotor, respectively.  The hyperparameters $\lambda_1, \lambda_2, \lambda_3, \lambda_4, \lambda_5$ trade-off objectives regarding perception awareness and thrust-to-weight ratio constraints against progress objectives.

\subsubsection{Policy Training}
All control policies are trained using Proximal Policy Optimization~(PPO)~\cite{schulman2017ppo}. 
PPO has been shown to achieve state-of-the-art performance on a set of continuous control tasks and is well suited for learning problems where interaction with the environment is fast.
Data collection is performed by simulating 100 agents in parallel using TensorFlow Agents~\cite{TFAgents}. 
At each environment reset, every agent is initialized in a random gate on the track layout with bounded perturbation around a state previously observed when passing the respective gate. 

\vspace*{-2pt}
\section{Experiments}
\vspace*{-2pt}
Using the training methodology described in the previous section, our experiments aim to answer the following research questions: 
(i)~Which of the architectures (Naive, Multihead, FiLM, our augmented FiLM) presented in the previous section (\ref{sec:conditioning}) is best suited for conditioning mobile robot control policies? 
(ii)~Is it better to use a discrete or continuous conditioning signal? 
(iii)~What role does the size of the network play? 
(iv)~Do the results transfer to a real-world quadrotor platform?

\vspace*{-2pt}
\subsection{Experimental Setup}
\vspace*{-2pt}
As an example for a mobile robot, this work uses the agile quadrotor platform shown in Fig.~\ref{fig:figure_1} with specifications listed in Table~\ref{tab:drone_specs}.
The evaluated control policies are all trained for the task of autonomous drone racing. 
In contrast to prior work tackling autonomous racing, our experiments focus on the racing performance when conditioned on a high-level user input. 
The user inputs evaluated in this work include (i)~constraints on the maximum agility while racing and (ii)~a user-defined perception objective.
While the former conditioning signal aims at steering the aggressiveness of the racing strategy and allows to trade off between speed and safety, the latter enables to alter the robot's heading direction during flight, which could be used to focus perception on salient landmarks on the track or keep an opponent in the field of view during a race.

Our study is performed on the three track layouts shown in Fig.~\ref{fig:racetracks} - a square track, a figure-8 track and a complex three-dimensional track layout~\cite{ackermann2022ieee} called \emph{Split-S} track, due to the maneuver required to pass the double gate on the far right.
Throughout all experiments, performance is measured by comparing the achieved laptimes and perception awareness of the deployed policies. 
Perception awareness is quantified by evaluating the average angular error between the desired and observed camera viewing direction.

\begin{table}[t!]
    \caption{\textnormal{Physical parameters of the quadrotor.}}
    \vspace*{-9pt}
    \label{tab:drone_specs}
    \centering
    \renewcommand{\arraystretch}{1.1}
    \begin{tabularx}{1.0\linewidth}{l|>$l<$|>$X<$}
    \toprule
    Parameter & \text{Unit} & \text{Value} \\
    \midrule
    Mass & \unit{kg} & 0.807 \\
    Thrust & \unit{N} & 36 \\
    TWR & \unit{-} & 4.5 \\
    Inertia & \unit{g\,m^2} & I_{xx} = 2.5,~~I_{yy} = 2.1,~~I_{zz} = 4.3 \\ 
    \bottomrule
    \end{tabularx}
    \vspace*{-12pt}
\end{table}

\vspace*{-2pt}
\subsection{Choice of Network Architecture}
\vspace*{-2pt}
\label{sec:netw_arch}
In a first set of simulation experiments we aim at identifying the best network architecture for efficient neural policy conditioning in autonomous drone racing. 
To this end, we focus on the most complex track layout \emph{Split-S} and measure the achieved laptime when conditioned on a maximum agility level. 
Specifically, the policies for all architectures are conditioned on the available thrust-to-weight ratio~(TWR), ranging from 1.6~TWR to 4.5~TWR. 
To have an estimate of the lower bound of the laptime, we also train so-called \emph{fixed-TWR} policies. 
These policies are trained for a single TWR setting, allowing them to overfit for a specific agility level, which typically results in faster training progress and superior performance.

Figure~\ref{fig:ablation_architecture} shows the results of this experiment. 
The fixed-speed reference is trained for and evaluated at 14 evenly spaced points throughout the TWR interval {$[1.6,~4.5]$}. 
Each of the conditioned policies are then evaluated at these thrust-to-weight ratio setpoints. 
To reduce the stochasticity of the results, each policy is the best of three identical policies trained with different initial random weights. 

\begin{figure}[t!]
    \centering
    \tikzstyle{every node}=[font=\footnotesize]
\definecolor{mycolor1}{rgb}{0.92900,0.69400,0.12500}%
\definecolor{mycolor2}{rgb}{0.85000,0.32500,0.09800}%
\definecolor{mycolor3}{rgb}{0.49400,0.18400,0.55600}%
\definecolor{mycolor4}{rgb}{0.46600,0.67400,0.18800}%
\definecolor{mycolor5}{rgb}{0.00000,0.44700,0.74100}%
\definecolor{mycolor6}{rgb}{0.30100,0.74500,0.93300}%
\begin{tikzpicture}

\begin{axis}[%
width=7.037cm,
height=3.7cm,
at={(0cm,0cm)},
scale only axis,
xmin=1.5000,
xmax=4.5000,
xlabel={Max. Thrust-to-Weight Ratio},
ymin=5.0000,
ymax=9.0000,
ylabel={Laptime [s]},
axis background/.style={fill=white},
xmajorgrids,
ymajorgrids,
legend style={legend cell align=left, align=left, draw=white!15!black},
legend columns=2,
xlabel shift = -0pt,
ylabel shift = -0pt,
line join = round,
scale only axis=true,
font = {\footnotesize},
title style = {font=\normalfont},
legend style = {font=\footnotesize},
legend pos = north east
]
\addplot [color=black, mark size=0.9pt, mark=square*, mark options={solid, fill=black, draw=black}]
  table[row sep=crcr]{%
1.5750	8.7137\\
1.8000	7.9727\\
2.0250	7.1577\\
2.2500	6.9863\\
2.4750	6.7264\\
2.7000	6.4608\\
2.9250	6.2626\\
3.1500	6.1971\\
3.3750	5.9015\\
3.6000	5.5964\\
3.8250	5.4326\\
4.0500	5.3461\\
4.2750	5.4018\\
4.5000	5.3461\\
};
\addlegendentry{Fixed TWR}

\addplot [color=mycolor1, mark size=2.2pt, mark=diamond*, mark options={solid, fill=mycolor1, draw=mycolor1}]
  table[row sep=crcr]{%
1.8000	8.0257\\
2.0250	7.3202\\
2.2500	7.1433\\
2.4750	6.8896\\
2.7000	6.6746\\
2.9250	6.4760\\
3.1500	6.3032\\
3.3750	6.0622\\
3.6000	5.7933\\
3.8250	5.5887\\
4.0500	5.5263\\
4.2750	5.5209\\
4.5000	5.5286\\
};
\addlegendentry{Naive-c}

\addplot [color=mycolor2, mark size=2.2pt, mark=diamond*, mark options={solid, fill=mycolor2, draw=mycolor2}]
  table[row sep=crcr]{%
1.8000	8.1679\\
2.0250	7.5537\\
2.2500	7.2823\\
2.4750	7.0048\\
2.7000	6.6958\\
2.9250	6.4929\\
3.1500	6.3075\\
3.3750	6.0614\\
3.6000	5.7737\\
3.8250	5.5521\\
4.0500	5.4338\\
4.2750	5.4334\\
4.5000	5.4334\\
};
\addlegendentry{Naive-d}

\addplot [color=mycolor3, mark size=0.8pt, mark=triangle*, mark options={solid, rotate=180, fill=mycolor3, draw=mycolor3}]
  table[row sep=crcr]{%
1.8000	8.2201\\
2.0250	7.4017\\
2.2500	7.1518\\
2.4750	6.8662\\
2.7000	6.6460\\
2.9250	6.5963\\
3.8250	5.6374\\
4.0500	5.5508\\
4.2750	5.5396\\
4.5000	5.5394\\
};
\addlegendentry{Multihead-d}

\addplot [color=mycolor4, mark size=1.2pt, mark=*, mark options={solid, fill=mycolor4, draw=mycolor4}]
  table[row sep=crcr]{%
1.5750	8.9926\\
1.8000	8.1089\\
2.0250	7.3791\\
2.2500	7.1874\\
2.4750	6.9399\\
2.7000	6.6961\\
2.9250	6.4652\\
3.1500	6.2766\\
3.3750	6.0427\\
3.6000	5.7836\\
3.8250	5.5634\\
4.0500	5.5248\\
4.2750	5.5242\\
4.5000	5.5189\\
};
\addlegendentry{FiLM-c}

\addplot [color=mycolor5, mark size=1.2pt, mark=*, mark options={solid, fill=mycolor5, draw=mycolor5}]
  table[row sep=crcr]{%
1.5750	8.7307\\
1.8000	7.8710\\
2.0250	7.1644\\
2.2500	6.9924\\
2.4750	6.7637\\
2.7000	6.5422\\
2.9250	6.3338\\
3.1500	6.1603\\
3.3750	5.9304\\
3.6000	5.6805\\
3.8250	5.4977\\
4.0500	5.4329\\
4.2750	5.4136\\
4.5000	5.4011\\
};
\addlegendentry{FiLM*-c}

\addplot [color=mycolor6, mark size=1.2pt, mark=*, mark options={solid, fill=mycolor6, draw=mycolor6}]
  table[row sep=crcr]{%
1.8000	8.1909\\
2.0250	7.4293\\
2.2500	7.1946\\
2.4750	6.9121\\
2.7000	6.6596\\
2.9250	6.4362\\
3.1500	6.2332\\
3.3750	6.0348\\
3.6000	5.7986\\
3.8250	5.5402\\
4.0500	5.4736\\
4.2750	5.4780\\
4.5000	5.4783\\
};
\addlegendentry{FiLM*-d}

\end{axis}
\end{tikzpicture}%
\tikzstyle{every node}=[font=\normalsize]
    
    \footnotesize
    \begin{tabularx}{1\linewidth}{X|cc}
    \toprule
    Architecture & Avg. Rel. Laptime [\%] & Max. Rel. Laptime [\%] \\
    \midrule
    Naive-c & 2.63 &	 3.52\\
    Naive-d & 3.25 &	 5.98\\
    Multihead-d & 3.23 &	 4.23\\
    FiLM-c & 2.80  &  3.64  \\
    FiLM*-c & \bf 0.54	& \bf 1.62 \\
    FiLM*-d & 3.82	& 4.69\\
    \bottomrule
    \end{tabularx}
    \caption{All architectures are able to condition a quadrotor control policy for agile flight on the maximum thrust-to-weight ratio. Our augmented \emph{FiLM*-c} architecture even manages to be within \unit[0.6]{\%} of a fixed-TWR baseline, indicating that one does not have to trade-off control performance for the added flexibility to regulate the controller during deployment. Furthermore, the \emph{FiLM}-based architectures cover the whole TWR-range and, unlike the \emph{Naive} baseline, do not crash at the lowest TWR setting.}
    \label{fig:ablation_architecture}
    \vspace*{-18pt}
\end{figure}

All the architectures we evaluated result in control policies that are able to race at a wide range of thrust-to-weight ratios. However, upon a closer look one can see that the \emph{FiLM*-c} policy leveraging our augmented FiLM architecture outperforms the other approaches in terms of laptime. More importantly, the \emph{FiLM*-c} is less than \unit[0.6]{\%} slower on average than a set of specifically trained fixed-TWR policies. We therefore gain the flexibility to regulate the neural controller during deployment while paying almost no penalty in terms of the optimality (i.e. laptime) of the control policy.
Furthermore, at the lowest thrust-to-weight setting of 1.6 TWR only the policy trained with the \emph{FiLM-c} and the \emph{FiLM*-c} architectures are able to race collision-free through the track. 
All other policies do not complete a single lap as they crash into the ground at some point. 
This further highlights the superior versatility of the FiLM architecture, as it is able to cover a wider range of conditioning inputs compared to the other architectures.
When comparing policies that operate on continuous inputs to policies trained using a one-hot encoding, we find that the continuous encoding outperforms its discrete counterpart both for the FiLM architecture as well as the naive architecture. 

Based on the results presented above, we conclude that the \emph{FiLM*-c} framework outperforms the other approaches and is extremely close to a fixed-TWR reference policy. We thus use this architecture in all subsequent experiments.

\vspace*{-2pt}
\subsection{Network Size}
\vspace*{-2pt}
We ablate the impact of changes in the network size on the performance of the \emph{FiLM*-c} policy. 
Both the policy-network and the value-network are implemented as two-layer MLPs and we vary their sizes together, such that the value network has four times wider layers. 
As in the previous experiments, all trained policies are conditioned on a maximum thrust-to-weight ratio while racing through the \emph{Split-S} track layout.

Consistent with the previous experiments, all policies are trained on a thrust-to-weight ratio interval of 1.6 to 4.5 for a fixed number of environment interactions. 
For each setting, three policies are trained to reduce the variance and all numbers are averages across those three policies.  
Table~\ref{tab:ablation_size} shows the average relative laptimes achieved by a \emph{FiLM*-c} controller with the different network sizes. 
One can see that a too small network is not expressive enough while larger network sizes become increasingly difficult to train in an RL-setting, indicated by the increased laptime.
Based on these results, we chose a \emph{FiLM*-c} architecture where the policy/value network have 128/512 neurons per layer.

\begin{table}[t!]
    \centering
    \caption{\textnormal{Average relative laptimes of a \emph{FiLM*-c} policy achieved when trained with \\[-4pt] different sizes of the policy- and value-network. }}
    \label{tab:ablation_size}
    \vspace*{-9pt} 
    \footnotesize
    \begin{tabularx}{1\linewidth}{X|cc}
    \toprule
    Network Size & Avg. Rel. Laptime [\%] & Max. Rel. Laptime [\%] \\
    \midrule
    64 & 5.14 & 13.26 \\
    \bf 128 & \bf 2.09 & 4.97 \\
    256 & 2.22 & \bf 3.2 \\
    512 & 3.87 & 4.76  \\
    \bottomrule
    \end{tabularx}
    \vspace*{-18pt} 
\end{table}

\vspace*{-2pt}
\subsection{Different Track Layouts}
\vspace*{-2pt}
After discussing the choice of network architecture, we now study how the selected architecture performs on the different track layouts introduced above (see Fig.~\ref{fig:racetracks}). 
We again consider the task of conditioning the policy on the maximally available thrust-to-weight ratio and summarize the results in Table~\ref{tab:ablation_tracks}.
The results obtained for the \emph{Split-S} and the \emph{Figure-8} track are very similar and verify that the \emph{FiLM*-c} architecture works on a variety of track layouts. 
On the \emph{Square} track, we obtain a surprising result: the \emph{FiLM*-c} architecture consistently outperforms the fixed-TWR reference. 
This result indicates that the policy is able to combine experience gained at different TWR settings and finds a general policy that is strictly better in terms of laptime than the individually trained fixed-TWR policies.

\begin{table}[b!]
    \centering
    \vspace*{-12pt}
    \footnotesize
    \caption{\textnormal{Comparison of the fixed of relative laptimes (w.r.t fixed-TWR reference)}}
    \vspace*{-9pt}
    \label{tab:ablation_tracks}
    \begin{tabularx}{1\linewidth}{X|cc}
    \toprule 
    & Avg. Rel. Laptime [\%] &  Max. Rel. Laptime [\%] \\
    \midrule
    Square Track & -4.60 & -3.39 \\
    Figure-8 Track & 0.50 & 3.14 \\
    Split-S Track & 0.54 & 1.62  \\
    \bottomrule
    \end{tabularx}
\end{table}

\vspace*{-2pt}
\subsection{Real-World Experiments}
\vspace*{-2pt}
The ablation studies presented in the previous sections were all conducted in simulation. 
In this section we present the transferability of our approach to real-world experiments conducted on an agile quadrotor platform. 
The platform used for these experiments is shown in \ref{fig:figure_1} and it matches our simulated quadrotor in specifications (see \ref{tab:drone_specs}). 
We encourage the reader to watch the supplementary video to understand the dynamic nature of these experiments. 

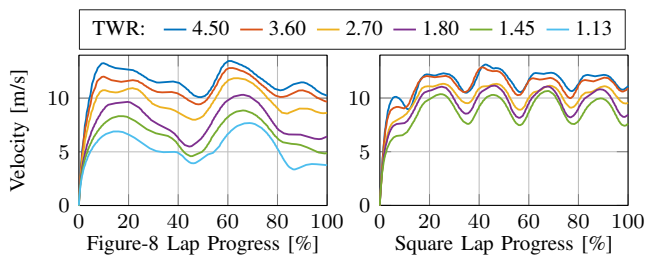
\begin{figure}[t!]
    \centering
    \vspace*{-6pt}
    \tikzstyle{every node}=[font=\footnotesize]
\definecolor{mycolor1}{rgb}{0.92900,0.69400,0.12500}%
\definecolor{mycolor2}{rgb}{0.00000,0.44700,0.74100}%
\begin{tikzpicture}

\begin{axis}[%
width=7.037cm,
height=3.2cm,
at={(0cm,0cm)},
scale only axis,
xmin=1.5000,
xmax=4.5000,
xlabel={Max. Thrust-to-Weight Ratio},
ymin=5.0000,
ymax=9.0000,
ylabel={Laptime [s]},
axis background/.style={fill=white},
xmajorgrids,
ymajorgrids,
legend style={legend cell align=left, align=left, draw=white!15!black},
legend columns=2,
xlabel shift = -3pt,
ylabel shift = -0pt,
line join = round,
scale only axis=true,
font = {\footnotesize},
title style = {font=\normalfont},
legend style = {font=\footnotesize},
legend pos = north east
]
\addplot [color=black, mark size=0.9pt, mark=square*, mark options={solid, fill=black, draw=black}]
  table[row sep=crcr]{%
1.8000	8.3749\\
2.2500	7.3187\\
2.7000	6.5047\\
3.6000	5.6746\\
4.0500	5.3876\\
4.5000	5.4284\\
};
\addlegendentry{Fixed TWR}

\addplot [color=mycolor1, mark size=2.2pt, mark=diamond*, mark options={solid, fill=mycolor1, draw=mycolor1}]
  table[row sep=crcr]{%
1.8000	8.7861\\
2.7000	7.1790\\
3.6000	5.7123\\
4.5000	5.8236\\
};
\addlegendentry{Naive-c}

\addplot [color=mycolor2, mark size=1.2pt, mark=*, mark options={solid, fill=mycolor2, draw=mycolor2}]
  table[row sep=crcr]{%
1.8000	8.5086\\
2.7000	6.9591\\
3.6000	5.6411\\
4.5000	5.5012\\
};
\addlegendentry{FiLM*-c}

\end{axis}
\end{tikzpicture}%
\tikzstyle{every node}=[font=\normalsize]
    \vspace*{3pt}
    
    \footnotesize
    \begin{tabularx}{1\linewidth}{X|cc}
    \toprule
    Architecture & Avg. Rel. Laptime [\%] & Max. Rel. Laptime [\%] \\
    \midrule
    Naive-c & 4.52 &	 10.43 \\
    FiLM*-c & \bf 1.91	& \bf 6.98 \\
    \bottomrule
    \end{tabularx}
    \caption{The plot and table compare the laptimes achieved by the \emph{Naive-c} and \emph{FiLM*-c} approach in real-world experiments on the \emph{Split-S} track. Similar to the simulation results, the \emph{FiLM*-c} architecture outperforms the naive baseline.}
    \label{fig:realworld_splits}
\end{figure}

In a first set of experiments we repeat a subset of the experiments presented in \ref{sec:netw_arch} and compare the \emph{FiLM*-c} architecture with both fixed thrust-to-weight ratio policies and the \emph{Naive-c} network (see Fig.~\ref{fig:realworld_splits}). 
Similar to what we observed in simulation, the conditioning with the \emph{FiLM*-c}  works well and it outperforms the naive baseline in terms of laptime while being within \unit[2]{\%} of the fixed-TWR reference.

\begin{figure}[t!]
    \centering
    \input{media/RealworldSquare}
    \vspace*{-10pt}
    
    \footnotesize
    \begin{tabularx}{1\linewidth}{l|C|C}
    \toprule 
    TWR & Figure-8 Laptime [s] & Square Laptime [s] \\
    \midrule
    4.50 &  2.93 & 3.22 \\
    3.60 &  3.10 & 3.26 \\
    2.70 &  3.50 & 3.27 \\
    1.80 &  4.59 & 4.31 \\
    1.45 &  5.44 & -- \\
    1.13 &  6.52 & -- \\
    \bottomrule 
    \end{tabularx}
    \caption{The plot and table compare for each track how a \emph{FiLM*-c} policy performs in terms of achieved speeds and laptime. Especially on the \emph{Figure-8} track the policy manages to fly the quadrotor with as little thrust margin (w.r.t hover) as 13\% up to 350\%.}
    \label{fig:realworld_other}
    \vspace*{-18pt}
\end{figure}

We also evaluate the chosen \emph{FiLM*-c} conditioning approach on the two other tracks and show the flown trajectories for various user-defined thrust-to-weight ratios in Fig.~\ref{fig:realworld_other}. 
On the simple \emph{Figure-8} track, the \emph{FiLM*-c} policy can handle thrust-to-weight ratios as low as 1.13. 
The two plots in Fig.~\ref{fig:realworld_other} illustrate the observed speeds. 
From the speed-plots one can also intuitively understand why the conditioning approach is very successful: when the speed is plotted over the lap-progress the curves for all speed levels exhibit very similar features. 
Thus the implicit assumption behind the FiLM framework\textemdash that the tasks are similar and related via some continuous transformation\textemdash holds.

\vspace*{-2pt}
\subsection{Viewing Direction}
\vspace*{-2pt}
To demonstrate the generalizability of our proposed method, we extend the experimental evaluation with an additional demonstrator for policy conditioning: perception.
Specifically, we condition the \textit{viewing direction} of the quadrotor while racing through the same track layouts as in the previous experiments. 
The ability to actively control perception is extremely useful for a vision-based robot, as it allows to maintain visibility with visual landmarks and as a result can substantially improve performance of state estimation. 
We analyze conditioning on the viewing direction both in simulation and on a real-world robot. 
To this end, we task the quadrotor to race through the \emph{Split-S} track, while maintaining a user-specified heading direction relative to the next gate to be passed.

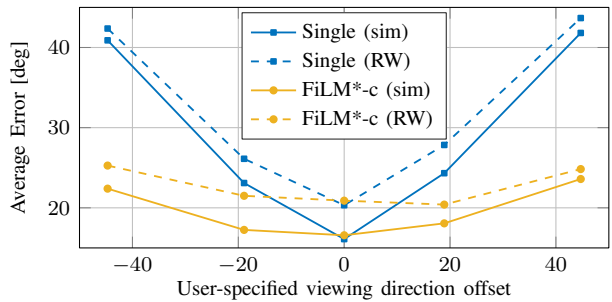
\begin{figure}[t!]
    \centering
    \tikzstyle{every node}=[font=\footnotesize]
\definecolor{mycolor1}{rgb}{0.00000,0.44700,0.74100}%
\definecolor{mycolor2}{rgb}{0.85000,0.32500,0.09800}%
\definecolor{mycolor3}{rgb}{0.92900,0.69400,0.12500}%
\definecolor{mycolor4}{rgb}{0.49400,0.18400,0.55600}%
\begin{tikzpicture}

\begin{axis}[%
width=7.037cm,
height=3.2cm,
at={(0cm,0cm)},
scale only axis,
xmin=-50.0000,
xmax=50.0000,
xlabel={User-specified viewing direction offset},
ymin=15.0000,
ymax=45.0000,
ylabel={Average Error [deg]},
axis background/.style={fill=white},
xmajorgrids,
ymajorgrids,
legend style={legend cell align=left, align=left, draw=white!15!black},
legend columns=1,
xlabel shift = -4pt,
ylabel shift = -0pt,
line join = round,
scale only axis=true,
font = {\footnotesize},
title style = {font=\normalfont},
legend style={font=\footnotesize, at={(0.5,0.98)},anchor=north}
]
\addplot [color=mycolor1, line width=0.7pt, mark size=0.9pt, mark=square*, mark options={solid, fill=mycolor1, mycolor1}]
  table[row sep=crcr]{%
-44.6907	40.8940\\
-18.9076	23.0940\\
0.0000	16.1020\\
18.9076	24.3220\\
44.6907	41.8120\\
};
\addlegendentry{Single (sim)}

\addplot [color=mycolor1, dashed, line width=0.7pt, mark size=0.9pt, mark=square*, mark options={solid, fill=mycolor1, mycolor1}]
  table[row sep=crcr]{%
-44.6907	42.3640\\
-18.9076	26.1190\\
0.0000	20.3270\\
18.9076	27.8430\\
44.6907	43.6720\\
};
\addlegendentry{Single (RW)}

\addplot [color=mycolor3, line width=0.7pt, mark size=1.2pt, mark=*, mark options={solid, fill=mycolor3, mycolor3}]
  table[row sep=crcr]{%
-44.6907	22.3930\\
-18.9076	17.2480\\
0.0000	16.5780\\
18.9076	18.0650\\
44.6907	23.5950\\
};
\addlegendentry{FiLM*-c (sim)}

\addplot [color=mycolor3, dashed, line width=0.7pt, mark size=1.2pt, mark=*, mark options={solid, fill=mycolor3, mycolor3}]
  table[row sep=crcr]{%
-44.6907	25.2710\\
-18.9076	21.5000\\
0.0000	20.8950\\
18.9076	20.4040\\
44.6907	24.8330\\
};
\addlegendentry{FiLM*-c (RW)}

\end{axis}
\end{tikzpicture}%
\tikzstyle{every node}=[font=\normalsize]
    \vspace*{-12pt}
    \caption{The \emph{FiLM*-c} architecture also generalizes to the task of conditioning a policy on the viewing direction (\emph{Split-S} track). A policy that is only trained for the single task to look at the next waypoint (Single) performs much worse than a policy that can be conditioned on the desired viewing offset (\emph{FiLM*-c}). This results holds both in simulation (sim) and real-world experiments (RW).}
    \label{fig:variable_perception}
    \vspace*{-18pt}
\end{figure}

Fig.~\ref{fig:variable_perception} shows the results of this experiment. 
Both in simulation and the real world, the proposed FiLM*-c approach maintains low heading errors over the entire spectrum of desired viewing directions. 
In contrast, policies that are trained for a single heading direction can not react to such user input and exhibit large errors for desired viewing directions different from zero.

\vspace*{-2pt}
\section{Conclusion}
\vspace*{-2pt}
This work presented a method to condition learning-based control policies for agile quadrotor flight on an auxiliary input. 
We evaluated different network architectures that process such user input through simple concatenation, multiple action heads, or by leveraging FiLM layers on the intermediate activations.
In an extensive ablation study, in simulation we compared the individual approaches by conditioning control policies on the maximally available thrust-to-weight ratio.
Our augmented FiLM architecture achieved the best performance and is less than \unit[0.6]{\%} (in simulation) or \unit[2]{\%} (in the real world) slower than a set of policies trained specifically for one thrust-to-weight ratio. 
When conditioning on the viewing direction offset w.r.t to the next landmark, there was no visible difference in laptime.
These findings implicate that we gain the additional flexibility to regulate a neural network controller and do not have to trade-off control performance. 
Therefore, we believe that this work is an important step in making neural controllers more accessible and safe to deploy for mobile robots.

\clearpage
\bibliographystyle{ieeetr}
\balance
\bibliography{references}

\begin{thebibliography}{10}

\bibitem{lee2020animalroughterrain}
J.~Lee, J.~Hwangbo, L.~Wellhausen, V.~Koltun, and M.~Hutter, ``Learning
  quadrupedal locomotion over challenging terrain,'' {\em Science robotics},
  vol.~5, no.~47, 2020.

\bibitem{tranzatto2021cerebus}
M.~Tranzatto, T.~Miki, M.~Dharmadhikari, L.~Bernreiter, M.~Kulkarni,
  F.~Mascarich, O.~Andersson, S.~Khattak, M.~Hutter, R.~Siegwart, and
  K.~Alexis, ``Cerberus in the darpa subterranean challenge,'' {\em Science
  Robotics}, vol.~7, no.~66, p.~eabp9742, 2022.

\bibitem{foehn2020alphapilot}
P.~Foehn, D.~Brescianini, E.~Kaufmann, T.~Cieslewski, M.~Gehrig, M.~Muglikar,
  and D.~Scaramuzza, ``Alphapilot: Autonomous drone racing,'' {\em Auton.
  Robots}, vol.~46, no.~1, p.~307–320, 2022.

\bibitem{ackermann2022ieee}
E.~Ackerman, ``{Autonomous Drones Challenge Human Champions in First "Fair"
  Race},'' {\em IEEE Spectrum}.

\bibitem{song2021autonomous}
Y.~Song, M.~Steinweg, E.~Kaufmann, and D.~Scaramuzza, ``Autonomous drone racing
  with deep reinforcement learning,'' {\em IEEE/RSJ Int. Conf. Intell. Robot.
  Syst. (IROS)}, 2021.

\bibitem{wurman2022gtsophy}
P.~R. Wurman, S.~Barrett, K.~Kawamoto, J.~MacGlashan, K.~Subramanian, T.~J.
  Walsh, R.~Capobianco, A.~Devlic, F.~Eckert, F.~Fuchs, L.~Gilpin,
  P.~Khandelwal, V.~Kompella, H.~Lin, P.~MacAlpine, D.~Oller, T.~Seno,
  C.~Sherstan, M.~D. Thomure, H.~Aghabozorgi, L.~Barrett, R.~Douglas,
  D.~Whitehead, P.~D{\"u}rr, P.~Stone, M.~Spranger, and H.~Kitano, ``Outracing
  champion gran turismo drivers with deep reinforcement learning,'' {\em
  Nature}, vol.~602, no.~7896, pp.~223--228, 2022.

\bibitem{romero2022mpcc}
A.~Romero, S.~Sun, P.~Foehn, and D.~Scaramuzza, ``Model predictive contouring
  control for time-optimal quadrotor flight,'' {\em IEEE Transactions on
  Robotics}, pp.~1--17, 2022.

\bibitem{perez2018film}
E.~Perez, F.~Strub, H.~de~Vries, V.~Dumoulin, and A.~Courville, ``Film: Visual
  reasoning with a general conditioning layer,'' {\em Proceedings of the AAAI
  Conference on Artificial Intelligence}, vol.~32, no.~1, 2018.

\bibitem{mees2022languageconditionedlearning}
O.~Mees, L.~Hermann, and W.~Burgard, ``What matters in language conditioned
  robotic imitation learning over unstructured data,'' {\em IEEE Robotics and
  Automation Letters (RA-L)}, vol.~7, no.~4, pp.~11205--11212, 2022.

\bibitem{lynch2020languageconditioned}
C.~Lynch and P.~Sermanet, ``Language conditioned imitation learning over
  unstructured data,'' {\em RSS: Robotics, Science, and Systems}, 2021.

\bibitem{reed2022gato}
S.~Reed, K.~Zolna, E.~Parisotto, S.~G. Colmenarejo, A.~Novikov, G.~Barth-maron,
  M.~Gim{\'e}nez, Y.~Sulsky, J.~Kay, J.~T. Springenberg, T.~Eccles, J.~Bruce,
  A.~Razavi, A.~Edwards, N.~Heess, Y.~Chen, R.~Hadsell, O.~Vinyals, M.~Bordbar,
  and N.~de~Freitas, ``A generalist agent,'' {\em Transactions on Machine
  Learning Research}, 2022.
\newblock Featured Certification.

\bibitem{codevilla2018endtoenddriving}
F.~Codevilla, M.~M{\"u}ller, A.~L{\'o}pez, V.~Koltun, and A.~Dosovitskiy,
  ``End-to-end driving via conditional imitation learning,'' in {\em 2018 IEEE
  international conference on robotics and automation (ICRA)}, pp.~4693--4700,
  IEEE, 2018.

\bibitem{blukis2018mappingnavigation}
V.~Blukis, D.~Misra, R.~A. Knepper, and Y.~Artzi, ``Mapping navigation
  instructions to continuous control actions with position-visitation
  prediction,'' in {\em Proceedings of The 2nd Conference on Robot Learning}
  (A.~Billard, A.~Dragan, J.~Peters, and J.~Morimoto, eds.), vol.~87 of {\em
  Proceedings of Machine Learning Research}, pp.~505--518, PMLR, 2018.

\bibitem{dosovitskiy2020yoto}
A.~Dosovitskiy and J.~Djolonga, ``You only train once: Loss-conditional
  training of deep networks,'' in {\em International Conference on Learning
  Representations}, 2020.

\bibitem{silva2021learningparametrizedskills}
B.~C. Da~Silva, G.~Konidaris, and A.~G. Barto, ``Learning parameterized
  skills,'' in {\em Proceedings of the 29th International Conference on
  International Conference on Machine Learning}, ICML'12, (Madison, WI, USA),
  p.~1443–1450, Omnipress, 2012.

\bibitem{deisenroth2014policysearch}
M.~P. Deisenroth, P.~Englert, J.~Peters, and D.~Fox, ``Multi-task policy search
  for robotics,'' in {\em 2014 IEEE International Conference on Robotics and
  Automation (ICRA)}, pp.~3876--3881, 2014.

\bibitem{dosovitskiy2017learning}
A.~Dosovitskiy and V.~Koltun, ``Learning to act by predicting the future,'' in
  {\em International Conference on Learning Representations}, 2017.

\bibitem{Abramson2021Deepmindmultimodal}
{DeepMind Interactive Agents Team}, J.~Abramson, A.~Ahuja, A.~Brussee,
  F.~Carnevale, M.~Cassin, F.~Fischer, P.~Georgiev, A.~Goldin, M.~Gupta,
  T.~Harley, F.~Hill, P.~C. Humphreys, A.~Hung, J.~Landon, T.~Lillicrap,
  H.~Merzic, A.~Muldal, A.~Santoro, G.~Scully, T.~von Glehn, G.~Wayne, N.~Wong,
  C.~Yan, and R.~Zhu, ``Creating multimodal interactive agents with imitation
  and self-supervised learning,'' 2021.

\bibitem{rahmatizadeh2018visionbasedmultitask}
R.~Rahmatizadeh, P.~Abolghasemi, L.~B\"{o}l\"{o}ni, and S.~Levine,
  ``Vision-based multi-task manipulation for inexpensive robots using
  end-to-end learning from demonstration,'' in {\em 2018 IEEE International
  Conference on Robotics and Automation (ICRA)}, p.~3758–3765, IEEE Press,
  2018.

\bibitem{schaul2015universalvalue}
T.~Schaul, D.~Horgan, K.~Gregor, and D.~Silver, ``Universal value function
  approximators,'' in {\em Proceedings of the 32nd International Conference on
  Machine Learning} (F.~Bach and D.~Blei, eds.), vol.~37 of {\em Proceedings of
  Machine Learning Research}, (Lille, France), pp.~1312--1320, PMLR, 2015.

\bibitem{jang2021bcz}
E.~Jang, A.~Irpan, M.~Khansari, D.~Kappler, F.~Ebert, C.~Lynch, S.~Levine, and
  C.~Finn, ``{BC}-z: Zero-shot task generalization with robotic imitation
  learning,'' in {\em 5th Annual Conference on Robot Learning}, 2021.

\bibitem{kaufmann2022benchmark}
E.~Kaufmann, L.~Bauersfeld, and D.~Scaramuzza, ``A benchmark comparison of
  learned control policies for agile quadrotor flight,'' in {\em 2022
  International Conference on Robotics and Automation (ICRA)}, IEEE, 2022.

\bibitem{shah2018airsim}
S.~Shah, D.~Dey, C.~Lovett, and A.~Kapoor, ``Airsim: High-fidelity visual and
  physical simulation for autonomous vehicles,'' in {\em Field and Service
  Robot.}, Springer, 2018.

\bibitem{furrer2016rotors}
F.~Furrer, M.~Burri, M.~Achtelik, and R.~Siegwart, ``Rotors—a modular gazebo
  mav simulator framework,'' in {\em Robot Operating System (ROS)}, Springer,
  2016.

\bibitem{foehn2022agilicious}
P.~Foehn, E.~Kaufmann, A.~Romero, R.~Penicka, S.~Sun, L.~Bauersfeld,
  T.~Laengle, G.~Cioffi, Y.~Song, A.~Loquercio, and D.~Scaramuzza,
  ``Agilicious: Open-source and open-hardware agile quadrotor for vision-based
  flight,'' {\em Science Robotics}, vol.~7, no.~67, 2022.

\bibitem{bauersfeld2021neurobem}
L.~Bauersfeld, E.~Kaufmann, P.~Foehn, S.~Sun, and D.~Scaramuzza, ``Neurobem:
  Hybrid aerodynamic quadrotor model,'' in {\em Proceedings of Robotics:
  Science and Systems}, 2021.

\bibitem{sun2019graybox}
S.~Sun, C.~C. de~Visser, and Q.~Chu, ``Quadrotor gray-box model identification
  from high-speed flight data,'' {\em Journal of Aircraft}, vol.~56, no.~2,
  pp.~645--661, 2019.

\bibitem{schulman2017ppo}
J.~Schulman, F.~Wolski, P.~Dhariwal, A.~Radford, and O.~Klimov, ``Proximal
  policy optimization algorithms,'' {\em ar{X}iv e-prints}, 2017.

\bibitem{zhou2019continuity}
Y.~Zhou, C.~Barnes, J.~Lu, J.~Yang, and H.~Li, ``On the continuity of rotation
  representations in neural networks,'' in {\em {IEEE} Int. Conf. Comput. Vis.
  Pattern Recog. (CVPR)}, 2019.

\bibitem{TFAgents}
S.~Guadarrama, A.~Korattikara, O.~Ramirez, P.~Castro, E.~Holly, S.~Fishman,
  K.~Wang, E.~Gonina, N.~Wu, E.~Kokiopoulou, L.~Sbaiz, J.~Smith, G.~Bartók,
  J.~Berent, C.~Harris, V.~Vanhoucke, and E.~Brevdo, ``{TF-Agents}: A library
  for reinforcement learning in tensorflow.''
  \url{https://github.com/tensorflow/agents}, 2018.
\newblock [Online; accessed 25-June-2019].

\end{thebibliography}

\end{document}